\documentclass[letterpaper, 10 pt, journal, twoside]{IEEEtran}
\usepackage{subcaption}
\usepackage{graphicx}
\usepackage{url}
\usepackage{verbatim}
\usepackage{cite}
\usepackage{times}
\usepackage{multicol}
\usepackage{multirow}
\usepackage[bookmarks=true]{hyperref}
\usepackage{ntheorem}
\newtheorem{hypothesis}{Hypothesis}

\begin{document}
\bstctlcite{IEEEexample:BSTcontrol}

\title{Robots Have Been Seen and Not Heard:\\ Effects of Consequential Sounds on Human-Perception of Robots}

\IEEEoverridecommandlockouts
\author{Aimee Allen$^{1}$, Tom Drummond$^{2}$ and Dana Kuli{\'c}$^{1}$%
    \thanks{Manuscript received: 2 August 2024; ~~Revised: 12 December 2024; Accepted: 4 February 2025.}
    \thanks{This paper was recommended for publication by Editor Angelika Peer upon evaluation of the Associate Editor and Reviewers' comments.
        This research was partly supported by an Australian Government Research Training Program (RTP) Scholarship. D. Kuli{\'c} is supported by the ARC Future Fellowship (FT200100761).}
    \thanks{$^{1}$A. Allen and D. Kuli{\'c} are with Faculty of Engineering, Monash University, Melbourne, Australia
    {\tt\footnotesize aimee.allen@monash.edu, dana.kulic@monash.edu}}%
    \thanks{$^{2}$T. Drummond is with School of Computing and Information Systems, University of Melbourne, Australia
    {\tt\footnotesize tom.drummond@unimelb.edu.au}}%
    \thanks{\copyright 2025 IEEE. Personal use of this material is permitted. Permission from IEEE must be obtained for all other uses, in any current or future media, including reprinting/republishing this material for advertising or promotional purposes, creating new collective works, for resale or redistribution to servers or lists, or reuse of any copyrighted component of this work in other works.}
    }

\maketitle

\begin{abstract}
Robots make compulsory machine sounds, known as `consequential sounds', as they move and operate. As robots become more prevalent in workplaces, homes and public spaces, understanding how sounds produced by robots affect human-perceptions of these robots is becoming increasingly important to creating positive human robot interactions (HRI). This paper presents the results from 182 participants (858 trials) investigating how human-perception of robots is changed by consequential sounds. In a between-participants study, participants in the sound condition were shown 5 videos of different robots and asked their opinions on the robots and the sounds they made. This was compared to participants in the control condition who viewed silent videos. Consequential sounds correlated with significantly more negative perceptions of robots, including increased negative `associated affects', feeling more distracted, and being less willing to colocate in a shared environment with robots. 
\end{abstract}


\section{Introduction} \label{sec:intro}
\IEEEPARstart{I}{t} is becoming increasingly common to find robots in public spaces, workplaces and even homes. Human perception of robots matters to successful human robot interactions (HRI) and adoption of robots in these shared spaces. A critical component influencing human perception is sound~\cite{Jouaiti2019}, a key element of human communication, and affecting mood and the ability to concentrate. However, many current robots can be quite loud or produce unpleasant sounds as they move, which can negatively impact human robot interaction (HRI). This paper studies how the sounds produced by robots as they move and operate are perceived by people, and thus the resulting implications for HRI and robot adoption in shared spaces.

\begin{figure}[t]
    \centering
    \includegraphics[width=0.9\linewidth]{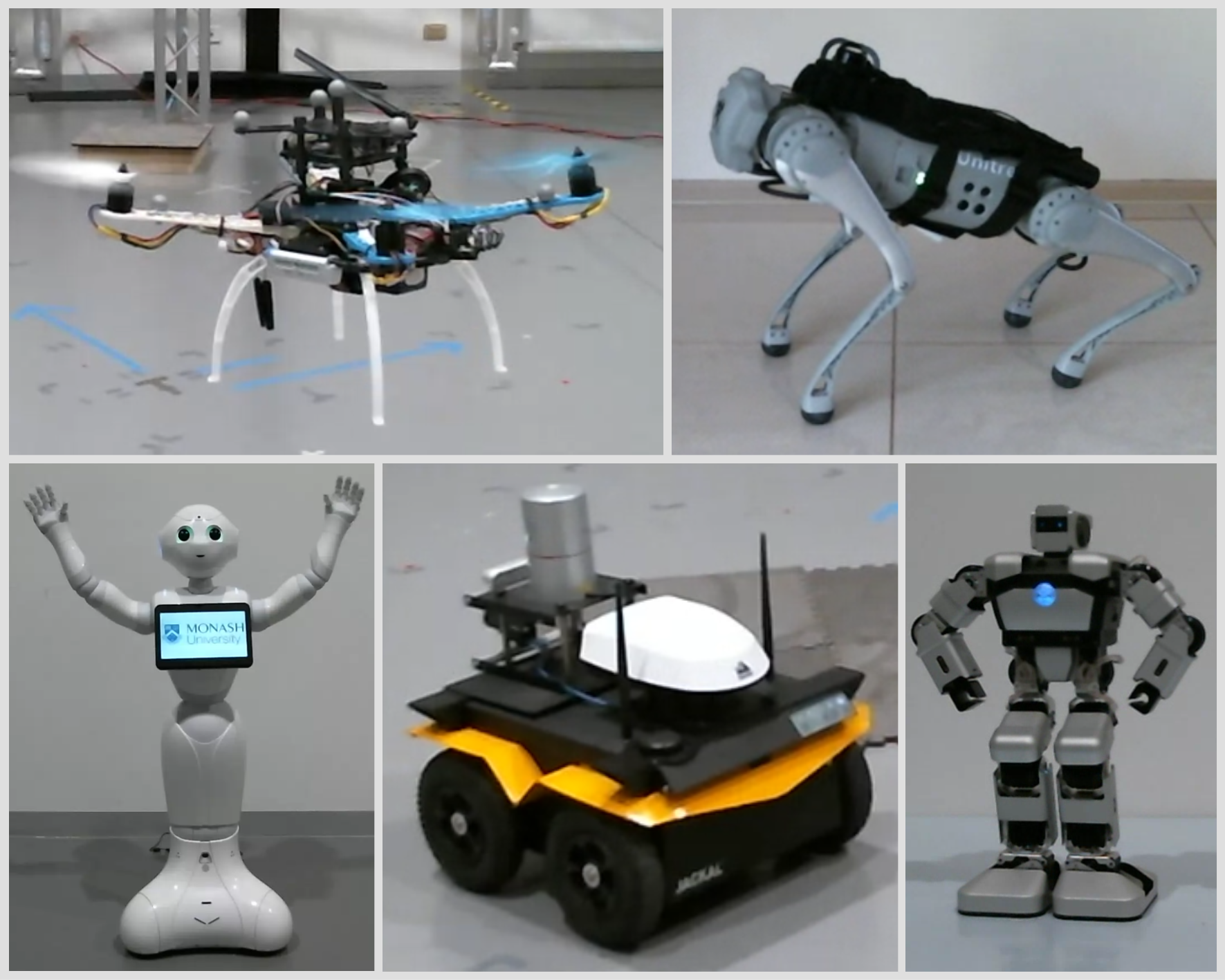}\hfill
    \caption{Robots featured in this study (top left to bottom right): Quadrotor; Go1 (Unitree); Pepper (SoftBank Robotics); Jackal (ClearPath Robotics); Yanshee (UBTECH).}
    \label{fig:five robots}
\end{figure}

Consequential sounds are the unintentional noises that a machine produces as it moves and operates\cite{Allen2023} i.e `sounds that are generated by the operating of the product itself'~\cite{Langeveld2013}. Robot consequential sounds are the audible `noises' produced by the actuators of the robot, and the robot can not function normally without making these sounds. Consequential sounds produced by a robot are influenced by many factors including types of actuators, robot form factor, and how the robot is moving~\cite{Wang2016,Allen2023}. Unpleasant machine generated sounds can have significant physiological effects on people including annoyance, anxiety, distraction and reduced productivity~\cite{DePaivaVianna2015,Basner2014,Jariwala2017}, with levels of auditory discomfort varying with individual differences in sensitivity and reactivity to sensory inputs\cite{Salvi2022,Henry2022,He2023}. However, to date, we do not have a detailed understanding of how robot consequential sounds influence human-perception of robots or human willingness to colocate with robots, necessary to inform  ways to manage, control or alter these sounds.

\noindent This paper makes the following contributions:
\begin{itemize}
    \item A study investigating the effects of consequential sounds produced by five different robotic platforms on human-perception of robots, focusing on key human-centric attributes needed for successful co-habitation: likeability, distraction by the robot, induced emotions, and willingness to colocate
    \item Quantitative analysis (N = 182) demonstrating that robot consequential sounds have a significant negative effect on human perception of the robots
\end{itemize}

\section{Related Work} \label{sec:related works}
\noindent Despite the importance of non-language sound in human-human interactions~\cite{Yilmazyildiz2016}, and the ubiquity of robot consequential sounds, they are rarely studied in HRI literature. A recent systematic review of non-verbal sound in HRI identified only seven relevant papers\cite{Zhang2023}.

\textbf{\emph{Psychoacoustics}:}
Sound is an important sensory element for humans~\cite{Jouaiti2019}, and the presence of sound during human-robot interaction can help localise robots within the environment and establish a sense of proxemic comfort~\cite{Cha2018,Trovato2018}. Sound can also have negative consequences, such as causing confusion or annoyance~\cite{DePaivaVianna2015,Schute2007,Moore2017}. As humans share environments with biological agents (such as other people or pets), they are accustomed to regular, rhythmic sounds such as breathing or rustling~\cite{Jouaiti2019}. Consequently, many people prefer sounds from embodied or moving agents to convey a sense of proxemics, i.e. a sound that denotes their presence and positioning~\cite{Trovato2018}. However, acute sounds (such as alarms) tend to disrupt focus by eliciting an instinctive danger or warning response~\cite{Jouaiti2019}. How people perceive and interpret auditory stimuli (including levels of sensitivity and reactivity) can vary between individuals, particularly for people with specific sound tolerance conditions (such as Hyperacusis\cite{Salvi2022}), noise sensitivities (such as Misophonia\cite{Henry2022}) or neurodiversities such as autism\cite{He2023}. Other influences such as culture may contribute to individual differences in sound perception, such as people from cultures with pitch sensitive languages who may be more affected by high or low pitched sounds\cite{Wong2012}. As humans are able to perceive small sound differences~\cite{Sneddon2003}, even minor changes in soundscape can have an impact on the perception of human robot interactions, which has been shown in existing research to influence opinions of particular robots~\cite{Moore2018,Frederiksen2019,Frederiksen2020,Cha2018,Song2017}.

\textbf{\emph{Consequential Sounds in HRI}:}
One of the first robotics papers to focus on ``consequential sound'', Moore et al.~\cite{Moore2017} compared videos of pairs of DC motors with artificially dubbed sounds, showing consistency of preferred sounds within participants, but not between participants. Another experiment overlaid consequential sounds from low quality robotic arms onto videos of a high quality KUKA robotic arm and found that both sets of consequential sounds reduced ratings, but with the higher quality robot sounds preferred~\cite{Tennent2017}. A more recent study investigated the effect of variance in sound intensity (volume) and frequency (pitch) of consequential sounds on perception of robots~\cite{Zhang2021} by manipulating consequential sounds in videos of a UR5 robot arm. Results suggested that quieter robots are less discomforting, and increasing sound frequency correlates with positive perceptions such as warmth. A study comparing the effectiveness of online HRI experiments looked at robot impressions for four conditions: in-person (with consequential sounds), soundproofed in-person, and two virtual conditions (both muted)~\cite{Izui2020}. This research found that the condition with consequential sounds showed worse robot impressions than the three (no sound) conditions. Several HRI experiments have shown that consequential sounds can negatively interact with other sounds or robot gestures, e.g low-frequency consequential sounds adding an unintended strong arousal, negative valence component which confused interpretation of intended affect~\cite{Frid2018,Frid2022} or participants choosing to limit robot movements to avoid generating disliked consequential sounds~\cite{Tian2021}.

Several research projects have investigated improving consequential sounds by adding transformative sounds on-top~\cite{Wang2023,Zhang2021b,Robinson2021,Frederiksen2019}. In ~\cite{Zhang2021b}, perception of warmth and competence were compared for several robots with and without the enhancements. In ~\cite{Wang2023}, birdsong and rain sounds were added to the consequential sounds of a micro-drone, showing that masked sounds were perceived as more pleasant than the unaltered sounds. Robinson et al.~\cite{Robinson2021} overlaid three carefully designed sounds on-top of the consequential sounds, and compared this to a silent control, however a comparison to perception of unedited consequential sounds was not made. Frederiksen et al.~\cite{Frederiksen2019} augmented consequential sounds by using headphones to attenuate existing sounds, and add affective sounds leading to the robot being perceived as happier, less angry and more curious. A recent work provides an in-depth explanation of consequential sounds within robotics~\cite{Allen2023}, including techniques for managing consequential sounds produced by robots.

Whilst existing research has helped to illustrate the effect that consequential sounds can have within human-robot interactions, very few studies have large samples, and all focus on robot-centric perception e.g. how competent or trustworthy the robot appears. This paper contributes critical human-centric perception elements such as how the robot makes the person feel, and willingness to colocate with robots, which are critical to understand for successful HRI.

\section{Research Methods} \label{sec:methods}
\noindent This paper aims to provide evidence that consequential sounds can create more negative human perceptions of robots and thus effect robot adoption. Two hypotheses are tested:

\begin{hypothesis}(H1) \label{hypothesis1}
    \textbf{Consequential sounds lead to more negative human perceptions of the robot} such as: negative feelings (e.g. being uncomfortable), distraction, less willingness to colocate with robots, and/or disliking the robot.
\end{hypothesis}

\begin{hypothesis}(H2) \label{hypothesis2}
    \textbf{Some robots are perceived more negatively than other robots due to their consequential sounds}.
\end{hypothesis}

\subsection{Experiment Protocol}
The experiment consisted of the following activities: a pre-questionnaire of demographic and robot-familiarity questions, sound tests to verify participants' ability to hear sounds, five randomly ordered trials each containing a 20 second video of a robot with per-trial questions, and a final post questionnaire\footnote{See research project website for robot videos and full questionnaires: \url{https://aimeeallen.github.io/perception-of-robot-consequential-sounds/}}. Participants were randomly assigned (50/50 condition split) between the consequential sound (CS) group, who saw five videos of robots with sound, and the control group of no-sound (NS), who saw visually identical but completely silent videos. The control condition provides a baseline reaction to visual elements such as appearance and movement of each robot, for a between-participant comparison on the influence of consequential sounds. Using a silent no-sound condition as the control (rather than a control with background noises only), provides consistency with prior work \cite{Robinson2021, Izui2020}. The experiment was run online using the Qualtrics platform and took participants roughly 30 minutes to complete. To mitigate limitations of the online format, sound tests were used, and contextual priming was provided to participants to match common attributes of many HRI deployments. In addition to where and how videos were filmed, context was provided to participants within the questionnaires by specifying colocation with robot (but not direct interaction), and within home, workplace and public-spaces. The study protocol was approved by Monash University Human Research Ethics Committee (MUHREC) (Project ID 36414). Participants were recruited via research connections, industry partnerships, social media, and snowballing with less than 3\% from survey-share platforms. The recruitment criteria were: over 18 years, access to a computer with internet and sound device, being able to respond to English questionnaires, and able to comfortably hear ambient room intensity sounds. Participants were not screened for or excluded based on any auditory impairment or sensitivity but may have chosen to self-exclude.

\subsection{Robot Videos}
Participants experienced five trials each consisting of a short 20 second robot video, followed by a questionnaire on that robot. Each video showed a single robot going through a large range of its typical motions that could be experienced if colocating with the robot, thus generating a variety of consequential sounds. As this study aims to model cohabitation (rather than direct interaction), robot actions and purpose were not explicitly explained to participants. This aligns with colocation scenarios where a robot's task may not be immediately decipherable by human co-inhabitants. Robots selected for this experiment were chosen to: a) be likely to exist within a variety of human-occupied environments (thus producing sound stimulus in human vicinity, potentially impacting human comfort, productivity and willingness to cohabitate), b) be diverse in terms of form factors, size, uses, motions and sounds produced, and c) include few enough robots/trials to not cause participant fatigue. Robot choice aimed for broad coverage within a HRI context, facilitating insights generally applicable to a wide variety of HRI scenarios (see figure~\ref{fig:five robots} for robots used in this study). The robots were filmed in settings with a typical sound profile (spectra and intensity) of a quiet office or home environment. The environment sound profile was kept consistent and quiet across the short 20 second interval to maintain audibility of the consequential sounds and limit distractions due to additional variance of sounds. Maintaining normal background sound context is important as environment sounds may alter perception by either amplifying responses or acting as a mask for the robot consequential sounds. The sound stimulus was captured following best practice recommendations for consequential sound recording and presentation~\cite{Allen2023} such as using an omnidirectional microphone capable of recording sounds across the full human audible spectrum placed near the participant.

\subsection{Questionnaires} \label{sec:questionnaires}
As no prior studies have measured these same elements of human-perception for sound, a custom questionnaire was developed based on existing standardised HRI scales including `Godspeed' safety questions\cite{Bartneck2009} and `Trust Perception Scale-HRI' prior experience with robots \cite{Schaefer2016}. The per-trial (robot) questionnaire contained Likert scale questions on aspects of human-perception of robots and follow-up exploratory qualitative questions. As it is very difficult to disentangle human perception of the sound of robots from other modalities such as appearance and movement of the robot, the experiment asked participants questions regarding a number of these other factors. Many of the questions were common between the sound condition and control groups, providing a baseline for analytically separating the effect of sound through between-participant testing, as both groups experienced all the same stimuli (except for the sounds). This paper focuses on the 11 Likert perception of robot questions which were shared between all trials and conditions. Each Likert question consisted of a 7-point scale ranging from negative (1) to positive (7) perception. The questions were grouped into 4 thematic scales representing a critical aspect of perception towards robots: `associated-affect' (anxious, agitated, unsafe, uncomfortable) induced by the robot (4 questions), `distracted' by the robot (1 question), desire to `colocate' (home, workplace, public space) with the robot (3 questions), and `like' (physical appearance, movements of, overall) the robot (3 questions). Before doing any robot trials, participants filled out a pre-questionnaire containing demographic information and questions on their current familiarity with robots. After the 5 trials, participants completed a post questionnaire to rank robots from least to most favourite (with reasons why), provided qualitative feedback on the sounds heard (sound condition only), and suggestions on how they want robots to sound in the future.

\subsection{Sound Controls}
Given the purpose of this experiment is to study natural human responses to consequential sounds, there is a risk of bias in responses if participants are consciously aware of this purpose, and presumably focus on listening to the sound. To reduce this bias, participants were initially not informed of the research purpose, and a broad range of questions were asked regarding the robot's appearance and movements as well as sound. Full disclosure of the research purpose was provided to participants during the final comparison questionnaire. Prior to viewing the robots, participants were given two brief sound tests to calibrate their responses to the sounds. These tests were explained to participants as necessary to ensure their sound device (headphones or speakers) were calibrated correctly. Participants were advised to use the sound device they felt most comfortable using when listening to music.

The first test checked that a participant's sound device was set to an appropriate volume to hear the sounds. Participants were played three different commonly identifiable broadspectrum sounds and asked to select what sound they heard from a list of seven options. One sound stimulus was entirely silent, and the other two were bird sounds and broadspectrum `noise'. Both the sound stimulus and options to select the sound heard were provided in a random order.
The second test contained a series of pure tones across the human audible spectrum (between 20Hz to 20kHz~\cite{Cha2018,Fastl2007}) which played at non equal intervals. Participants were asked to press a button each time they heard any sound, which identified which frequencies each participant was capable of hearing. Sound tests were done for both the sound and no sound (control) groups as people with reduced hearing quality may rely on other senses, thus influencing their perception of the robots in other ways. Participants who failed these tests (2\%) were removed from the analysis.

\section{Results} \label{sec:results}
\noindent Results are reported following the guidelines of Hoffman\cite{Hoffman2020}. Firstly cohort characteristics are presented. Secondly, quantitative human-perception questions are combined into 4 scales and checked for internal consistency. Thirdly, regression analysis with `participant condition'(CS or NS) and `robot' predictor variables was used to model the effect on each of the 4 scales to evaluate Hypotheses 1 and 2.

\subsection{Cohort Characteristics}
\noindent This study recruited 186 total participants. Most participants completed the full five trials, leading to 878 completed trials. Results of the first sound calibration test were analysed for verification that participants could hear sound stimuli. All three questions were answered correctly by 162 participants (87\%), with 20 participants (11\%) getting one question wrong. The four participants (2\%) who got one or no questions correct (thus failing the sound test) were removed from the data set, leaving 182 participants and 858 trials. The 858 completed trials (across all robots) allow the detection of an effect size of $f^2=0.02$ with .91 power at an alpha level of .05 (calculated using the G*Power software\cite{GPowerFaul2007}.) Participant numbers in each condition were fairly even with 48.9\% experiencing the sound condition. Trial numbers for each robot-condition were similar, with 80-84 (sound condition) and 89-91 (control) trials each. Summary cohort statistics show good diversity within the sample population, e.g. age groups: 18 - 24 (13.7\%), 25 - 34 (19.8\%), 35 - 44 (19.8\%), 45 - 54 (12.1\%), 55 - 64 (11.5\%), 65+ (23.1\%), and gender: female (54.4\%), male (43.4\%), other (2.2\%). There was also good diversity across robot exposure frequency: Daily (19.8\%), Weekly (21.4\%), Monthly (14.8\%), Less Frequently (22.0\%), and Almost Never (22.0\%).

\subsection{Correlation between Question Scales}
\begin{figure}
    \centering
    \includegraphics[width=0.94\linewidth]{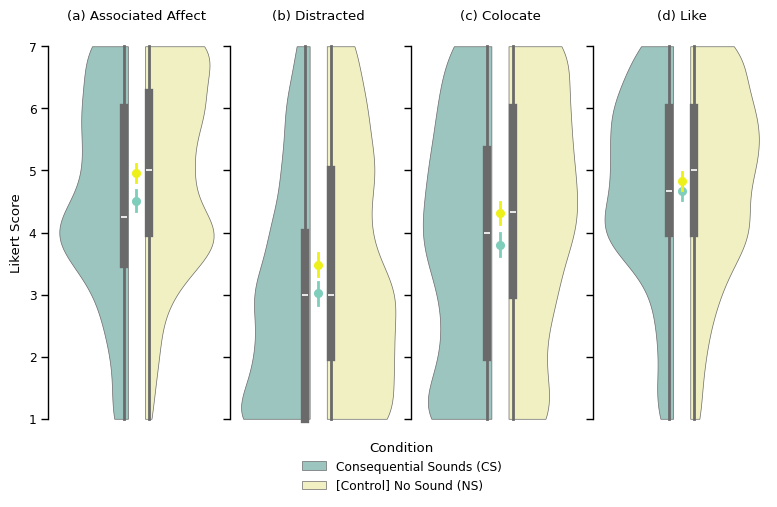}\hfill
    \caption{Data distributions between conditions for all 4 scales, (1) = negative to (7) = positive perception. Means (coloured dots) and 95\% confidence intervals (vertical lines) are shown between conditions.}
    \label{fig:plot-cond-only}
\end{figure}

\noindent The 11 per-trial Likert questions were grouped into 4 related question scales (see section~\ref{sec:questionnaires}). A Cronbach's alpha analysis was done to evaluate the internal consistency of the non-standardised questionnaires. All alphas were Good (colocate = .885) or Excellent (associated affect = .961, like = .916), with small confidence intervals (less than .03 range). As the scale correlations were strong, the 4 scales were used for all analysis. A combined value for responses to questions within the same scale was computed by treating the Likert values as continuous variables and averaging the responses to each question e.g. participant responses of (3, 7, 4, 3) for `associated-affect' questions, results in a 4.25 scale value. Comparing between the two conditions (CS and NS), the means for each scale are: Associated Affect $M_{CS}=4.51$, $M_{NS}=4.95$, Distracted $M_{CS}=3.03$, $M_{NS}=3.48$, Colocate $M_{CS}=3.80$, $M_{NS}=4.32$ and Like $M_{CS}=4.67$, $M_{NS}=4.83$ (see figure~\ref{fig:plot-cond-only}).

\subsection{Regression Analysis} \label{sec:results-regression}
\noindent To test H1 and H2, least-squares linear regression was performed for the 4 human-perception of robots scales (dependent variables) using `participant condition', `robot' and `interactions between them' as predictors. We follow the recommendation of Hoffman et al.~\cite{Hoffman2020} and apply regression analysis, as it provides clear identification of significant predictor terms for all predictor types (e.g. categorical or ordinal), provides flexibility for coding methods (to more intuitively relate to hypotheses), and allows changing predictors during analysis (removing non-significant predictors or post-hoc exploration of new predictors).

\begin{table}
    \centering
    \caption{Regression results for 4 human-perception scales. *Yanshee coefficients calculated from the other robots.}
    \label{tab:regression}

    \caption*{Associated Affect $R^2(adj) = .120, F(5, 826) = 23.56$}
    \begin{tabular}{|p{0.47\linewidth}|p{0.09\linewidth}|p{0.07\linewidth}|p{0.10\linewidth}|}
		\hline
		\textbf{Associated Affect (from robot)}&$\beta$&\textit{t}&\textit{p}-value\\\hline\hline
		Mean (for this scale)&4.956&68.69&-\\\hline
        Robot: Go1&-0.510&-4.83&$<$.001\\\hline
        Robot: Jackal&0.241&2.31&.021\\\hline
        Robot: Pepper&0.524&5.00&$<$.001\\\hline
        Robot: Quadrotor&-0.740&-7.09&$<$.001\\\hline
        Robot: Yanshee&0.485*&NA&NA\\\hline
        \textbf{Condition: Consequential Sounds}&-0.448&-4.27&$<$.001\\\hline
    \end{tabular}
    \bigskip
    
    \caption*{Distracted $R^2(adj) = .140, F(5, 848) = 28.68$}
    \begin{tabular}{|p{0.47\linewidth}|p{0.09\linewidth}|p{0.07\linewidth}|p{0.10\linewidth}|}
		\hline
		\textbf{Distracted (by robot)}&$\beta$&\textit{t}&\textit{p}-value\\\hline\hline
		Mean (for this scale)&3.480&42.32&-\\\hline
        Robot: Go1&-0.651&-5.44&$<$.001\\\hline
        Robot: Jackal&0.308&2.59&.010\\\hline
        Robot: Pepper&1.014&8.50&$<$.001\\\hline
        Robot: Quadrotor&-0.848&-7.12&$<$.001\\\hline
        Robot: Yanshee&0.177*&NA&NA\\\hline
        \textbf{Condition: Consequential Sounds}&-0.458&-3.84&$<$.001\\\hline
    \end{tabular}
    \bigskip

    \caption*{Colocate $R^2(adj) = .075, F(5, 823) = 14.52$}
    \begin{tabular}{|p{0.47\linewidth}|p{0.09\linewidth}|p{0.07\linewidth}|p{0.10\linewidth}|}
		\hline
		\textbf{Colocate (with robot)}&$\beta$&\textit{t}&\textit{p}-value\\\hline\hline
		Mean (for this scale)&4.323&48.02&-\\\hline
        Robot: Go1&-0.122&-0.94&.348\\\hline
        Robot: Jackal&0.037&0.28&.776\\\hline
        Robot: Pepper&0.589&4.51&$<$.001\\\hline
        Robot: Quadrotor&-0.850&-6.53&$<$.001\\\hline
        Robot: Yanshee&0.346*&NA&NA\\\hline
        \textbf{Condition: Consequential Sounds}&-0.525&-4.04&$<$.001\\\hline
    \end{tabular}
    \bigskip

    \caption*{Like $R^2(adj) = .037, F(5, 841) = 7.45$}
    \begin{tabular}{|p{0.47\linewidth}|p{0.09\linewidth}|p{0.07\linewidth}|p{0.10\linewidth}|}
		\hline
		\textbf{Like (robot)}&$\beta$&\textit{t}&\textit{p}-value\\\hline\hline
		Mean (for this scale)&4.827&67.89&-\\\hline
        Robot: Go1&-0.267&-2.59&.010\\\hline
        Robot: Jackal&0.051&0.50&.621\\\hline
        Robot: Pepper&0.428&4.16&$<$.001\\\hline
        Robot: Quadrotor&-0.410&-3.96&$<$.001\\\hline
        Robot: Yanshee&0.198*&NA&NA\\\hline
        \textbf{Condition: Consequential Sounds}&-0.164&-1.59&.112\\\hline
    \end{tabular}
\end{table}

The categorical `robot' variable was coded for the regression using deviation coding, which allows the mean for each robot to be compared to the overall mean across all robots (rather than to a specific reference robot). The coefficient for each robot $\beta_r$ becomes an offset from the overall mean across all robots.

Hypothesis 2 considers whether some robots have a stronger effect on the perception of sound than others. To address this, 3 terms need to be considered, condition (sounds heard or not), robot, and the interaction between these two. The regression is expressed as:

\begin{equation}
y_i = \beta_0 + \beta_c x_{ci} + \sum\limits_{r\in Robots} \beta_r x_{ri} + \sum\limits_{r\in Robots} \beta_{cr} x_{ci} x_{ri} + \varepsilon_i
\end{equation}

\noindent where $y_i$ = predicted human-perception scale value, $i$ = data point, $c$ = condition (sound), $r$ = robot, $\beta_0$ = mean of all data for the scale, $\varepsilon_i$ = residual error. Python libraries `patsy' and `statsmodels' were used to compute the coding design matrix and run the regression respectively. This regression showed that almost all interaction effects between the two predictors were non-significant $p\in[.194,.934]$, with two borderline significant effects for `Associated Affect' (Quadrotor with sound) ($\beta=-0.413,t=-1.98,p=.048$) and `Distracted' (Pepper with sound) ($\beta=0.447,t=1.87,p=.061$). As interactions were not significant, these were removed from the regression equation following best practices to improve precision of the model.

The revised regression includes only the robot type (independent of sound) and separate consequential sound condition, i.e. $\beta_{cr} = 0$. The results of this regression are shown in table \ref{tab:regression}. In the table, the significant `Condition: Consequential Sounds' lines represent how much lower (-ve coefficient) robots were rated when sound could be heard, addressing hypothesis H1. The significant robot values show that the robots are perceived differently from each other, but independent of sound.

A post-hoc analysis was done to examine the effect of `robot exposure frequency' on the 4 human-perception scales. Robot exposure was binned into two categories, `frequent exposure' (daily, monthly, weekly) versus `limited exposure' (less frequent, almost never) with similar and even split between the two sound conditions: CS (52\% frequent, 48\% limited), NS (60\% frequent, 40\% limited). A supplementary regression adding terms for robot exposure, and interaction effects was run with no new terms found to be significant for Associated Affect or Distracted scales. For the Collocate scale, `robot exposure frequency' was positively correlated ($\beta = 0.78, t = 4.29, p <.001$) but interaction with the sound or robot was not significant. For the Like scale, exposure was positively correlated ($\beta = 0.62, t = 4.33, p < .001$), and the interaction of exposure and sound was also negatively correlated ($\beta = -0.51, t = -2.48, p = .013$).

\subsection{Analysis by Scale} \label{sec:analysis-by-scale}

\begin{figure}
    \centering
    \includegraphics[width=0.91\linewidth]{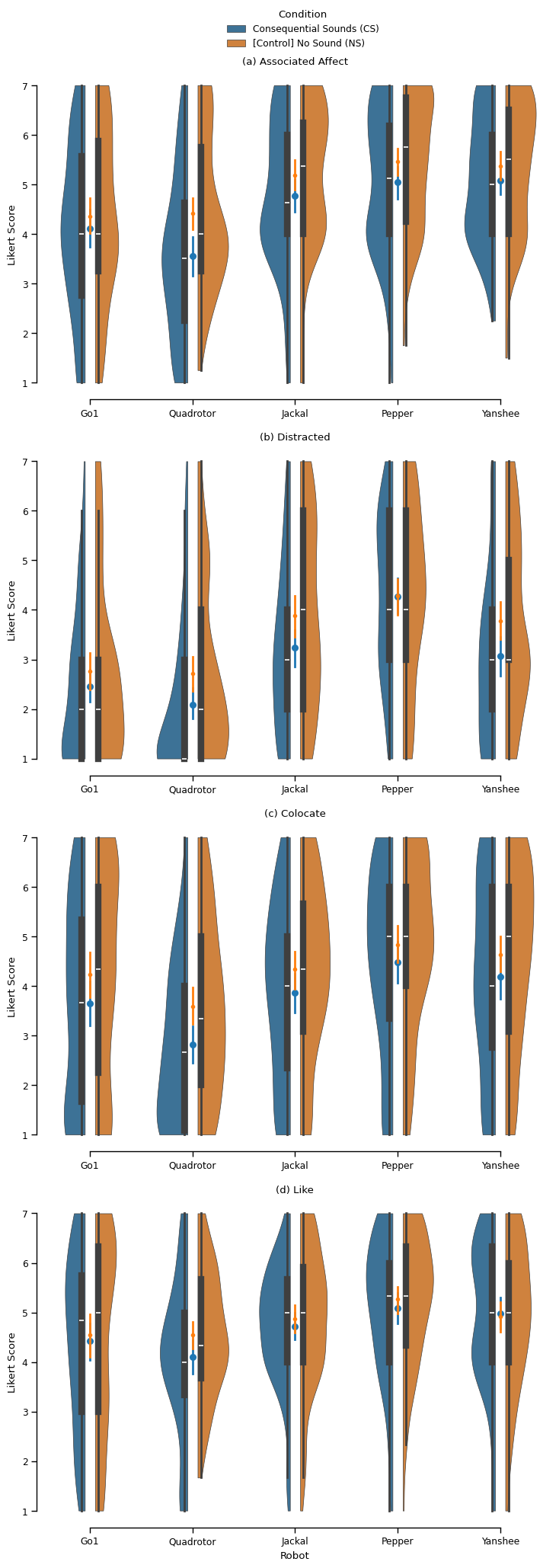}\hfill
    \caption{Data distributions across robots and conditions for all 4 scales, (1) = negative to (7) = positive perception. Means (coloured dots) and 95\% confidence intervals (vertical lines) are shown between condition pairs.}
    \label{fig:plot-robot-cond}
\end{figure}

Visualisations of the experiment data across the 4 human-perception scales can be seen in figures~\ref{fig:plot-cond-only} (split by condition) and \ref{fig:plot-robot-cond} (condition and robot), with regression results in table~\ref{tab:regression}. These representations provide the following insights.

\subsubsection{\textbf{Associated Affect}}
In the `Associated Affect’ scale, participants were asked how anxious, agitated, unsafe and uncomfortable the robot made them feel. The regression results (and figure~\ref{fig:plot-cond-only}(a)) show that the effect of sound was statistically significant. These results suggest that consequential sound contributes a strong negative effect as to how robots are perceived on the associated affect scale.

Considering the interaction between different robots and consequential sounds, the regression including interaction effects for `associated affect' had non-significant \textit{p}-values for all robot-condition interactions, with the exception of the Quadrotor which showed borderline significance, also shown visually in figure~\ref{fig:plot-robot-cond}(a). Whilst it would require further research to confirm, this suggests that the presence of Quadrotor (consistent rotary) consequential sounds may be more negatively perceived than other robot sounds.

Many participants additionally provided qualitative feedback on how consequential sounds influenced their affects such as, ``the quieter movement sounds are better, they are less distracting and make me feel calmer. The loud noises made me anxious, and I couldn't appreciate the tech in the robot because of it.'', and ``The robot makes a loud uncomfortable noise. I personally don't like it, and it makes me feel a bit anxious".

\subsubsection{\textbf{Distracted}}
The `distracted' by the robot scale asked participants how much they thought the robot would distract or interrupt them on a scale of `very distracted by the robot'(1) to `able to focus as if the robot wasn't there' (7). The regression results show a statistically significant correlation between participants being distracted and sound, as seen in figure~\ref{fig:plot-cond-only}(b). These results suggest that consequential sound contributes a strong negative effect as to how distracted participants feel. No interactions were significant, see figure~\ref{fig:plot-robot-cond}(b). Interestingly, the `distract' scores are very negative across all robots, which suggests people may think all robots are distracting to be around.

Additional qualitative feedback was provided by many participants as to how hearing consequential sounds influenced how distracted they felt, such as ``Clanky plastic noise from rapid movement is distracting.'' and ``(the robot) was very loud and distracting. Perhaps the sudden short noises were the most annoying.''

\subsubsection{\textbf{Colocate}}
The `colocate' scale asked participants if they would enjoy having the robot in their home, workplace or a shared public space. The regression results (table~\ref{tab:regression}) show a statistically significant negative effect from consequential sound on willingness to colocate, as seen in figure~\ref{fig:plot-cond-only}(c). 

Many participants shared their views on colocating with the robots such as ``(the robot is) just too noisy. Imagine owning a few of these at home, or at work. Headache central due to sounds" and ``the noise in the movement would make me not as interested in having (the robot) around''.

\subsubsection{\textbf{Like}}
The `like' robot scale asked participants how much they liked attributes of the robot such as its appearance, motions and overall likability. The regression results (table~\ref{tab:regression}) show a non-significant effect of sound for liking robots, illustrated in figure~\ref{fig:plot-cond-only}(d). These results contain insufficient evidence (given the current sample size) to conclude that consequential sounds impact how much a robot is liked. Further research is needed to investigate the effect of consequential sounds on how much a robot is liked.

\section{Discussion and Future Work}
\noindent \textbf{Supported hypotheses:} Consequential sound negatively impacts associated affects, ability to focus (people are more distracted) and willingness to colocate with robots, in support of H1. H2 suggests that different robots will have a more extreme effect on human-perception due to their consequential sounds. When analysing the interaction effects between robot and condition across the 4 scales, there was only a single robot in the `associated affect' scale which showed some statistical significance, thus there is insufficient evidence to support hypothesis 2. Further research is required to understand the effects of consequential sounds at an individual robot level.

\textbf{Connection to existing literature: }
The research presented in this paper is consistent with previous findings in related literature, that consequential sounds have a negative impact on elements of HRI due to how the sounds are perceived~\cite{Frid2018,Jouaiti2019,Trovato2018}. Whilst existing literature focuses on the perception of attributes of the robots e.g. was the robot happy, expensive, competent or dangerous~\cite{Izui2020,Moore2017,Robinson2021,Zhang2021}, this paper contributes evidence on attributes of the human co-inhabitants e.g. how the human felt and whether they wanted to colocate with the robot.

\textbf{Managing consequential sounds: }
As consequential sounds exist for all robots, this prompts the need to consider how consequential sounds can be attenuated, changed or removed to mitigate negative perceptions, such that people willingly accept robots into shared spaces. The first step to addressing consequential sounds is understanding which sound structures, both objective (power, frequencies) and subjective (acute sounds, pleasantness, noisiness) are causing negative perceptions (an area deserving of more research), before developing techniques to address these. Options to reduce the effects of consequential sounds include removal (such as dampening or cancellation), and augmentation (such as sound transformation or masking). Hardware sound dampening techniques are well known, however they may not be suitable for robot deployments as they can be expensive, typically only address component level sounds (potentially missing sounds related to the combination of components), or may create compromises on robot performance. Sound cancellation techniques may be hard to implement in large or unknown physical spaces, and may only be effective for people placed in specific locations relative to the cancellation. Software augmentation techniques (such as adding sounds to hide `bad' sounds thus altering human perception), show good promise, as they are adaptable in real-time to context and individuals. Further suggestions for managing robot consequential sounds in HRI research and robot design are provided in recent literature~\cite{Allen2023}.

\textbf{Limitations: }
The study presented in this paper has some limitations. Firstly, only 5 specific robots were tested, so the above results may not hold true for all robots. Similarly, the short time interval of videos meant it was not possible to include all possible movements (and thus sounds) that the robots could generate. For example, movement of Pepper's base would have added some additional sounds to the spectrum in the trial stimuli. Secondly, the study was conducted online via video-based stimuli, which (whilst a common approach in HRI) may provide less salient audio and context cues compared to an in-person study. The study results may not fully capture how humans perceive and respond to robots in their real world surroundings. Choice of auditory device (speakers versus headphones) can also influence listeners' affective perception, attitudes, and physiological responses to sound stimuli~\cite{Lieberman2022,Kallinen2007}. Listening via headphones may increase feelings of closeness to the sound source (as if inside their own head), which may exaggerate affective responses. Regarding video-stimulus volume, sound tests provided an initial calibration volume, and many participants would not self-select to listen to the sounds at an uncomfortable volume, however it was not possible to prevent participants choosing to change sound volume during the study, or finding specific sound profiles uncomfortable. The results should thus be interpreted acknowledging these differences and limitations of an online video-stimulus study compared to an in-person study, with prior HRI research suggesting it is likely that future real-world results may show an even stronger negative correlation~\cite{Jouaiti2019,DePaivaVianna2015,Izui2020}. Thirdly, context of an interaction is often important to how well robots are perceived~\cite{DePaivaVianna2015,Schute2007,Allen2023}. This research investigated user's opinions of location contexts including home, workplace and public spaces. A quiet and consistent background sound profile was used for the short videos, which may not accurately represent the diverse and dynamic contexts where robots could be used. The silent condition videos did not contain this background sound, which may have introduced some bias in human-perception between the conditions. Examining more environmental sound variations across a longer time interval, and considering other contexts such as different human activities (e.g. cognitive tasks, relaxing or socialising) would further support generalizability of these results. Finally, this research only included 4 elements of human-centric perception of robots, with findings derived from non-standardised questionnaires. It is possible that other related aspects of human perception may be affected in different or similar ways.

\textbf{Future Work: }
Future work should include in-person verification to address some aforementioned limitations by enabling investigation of how proxemics and perception of sound in a 3D space influence human perception. Prior research suggests that in-person exposure to consequential sounds correlated with strong negative responses~\cite{Izui2020, Tian2021}, so in-person studies may contribute stronger significance of results. In addition to a completely silent control condition, future HRI studies could benefit from including background environment sounds (without robot sounds) as additional comparative sound conditions against the robot consequential sound stimuli. Whilst including a second control comes with the risk of having too many comparative conditions, it could also strengthen future research by facilitating investigation into how environmental sounds interact with robot consequential sounds to affect human-perception, and address possible biases between silent and full-sound conditions.

Further exploratory research could uncover correlations between negative perceptions and specific groups of people e.g. introverts vs extraverts, alongside individual differences in perceptions and adverse side effects from hearing robot sounds, particularly for those individuals with auditory related conditions. Direct robot interaction versus only cohabitation should also be investigated, alongside the effect of what actions a robot is taking, and the co-inhabitant's knowledge of the purpose (and thus perceived usefulness) of the robot generating those actions and sounds. In-depth audio analyses could be done to uncover specific features of consequential sounds which may be perceived negatively (e.g. intensity (loudness), frequency, or abruptness of a sound), providing sound properties to target for improved human robot interaction. Commonalities across robot sounds in this study can be observed in the videos or via spectrograms\footnote{See research project website (provided in footnote 1)}. Many of the robots have more power (volume) in the higher frequency ranges whilst the robot is moving, with the Quadrotor and Go1 also showing more power in the lowest frequency bands. Several robots generate acute sound spikes (known to be perceived poorly by humans~\cite{Jouaiti2019,Trovato2018}) as they move. The Quadrotor has distinct harmonic bands of buzzing from its rotors. These observations correlate with many of the qualitative participant responses. Further research should also examine how people want robots to sound, for example many participants in this study suggested they wanted less noisy, lower frequencies and more organic (natural or animal-like) sounds for robots.

The effect of prior experience with robots warrants further study. The negative correlation between the interaction of robot experience with sound, and disliking a robot (presented in section~\ref{sec:results-regression}) suggests that people who are familiar with robots may be more likely to relate disliked sounds to prior longer experiences, which may have implications for whether people habituate to robot consequential sounds. Additional longer time-frame studies should be done to examine whether the negative perceptions associated with robotic consequential sounds are due to their unfamiliar nature (and may be potentiality habituated or sensitised to) or from specific sound properties. As there are currently no good models for predicting habituation (decreased response over time) versus sensitisation (increased response to stimulus) prior to exposure\cite{Blumstein2016}, understanding this could facilitate planning robot deployments to intentionally habituate people to robot sounds thus reducing negative perceptions over time. Importantly, more research is needed on methods for targeting and improving consequential sounds (e.g., sound augmentation) to reduce the negative human-perception towards robots.

\section{Conclusions} \label{sec:conclusion}
\noindent As machines, robots are forced to make certain sounds as they operate. This paper examined the effect of these `consequential sounds' on human perception of robots. Participants watched videos of 5 robots as they moved and produced consequential sounds, compared to participants in a silent control condition. Four elements of perception were assessed, `liking' the robot, `affects' induced by the robot (e.g. safety or comfort), how much the robot `distracted' a participant, and willingness to `colocate' with the robot in their home, workplace or a public space. Consequential sounds were found to significantly affect how people perceive the robots, leading to people feeling more distracted, experiencing stronger negative affects, and reducing desire to colocate with robots. The presence of sound showed no significant effect on how much the robots were liked. These significant negative human-perceptions of robots are likely to have an unwanted effect on robot adoption in shared spaces. Future work should identify which specific elements of consequential sounds cause these negative perceptions, alongside techniques to improve the sounds robots make to negate the effects of consequential sounds.

\bibliographystyle{IEEEtran}
\bibliography{arxiv}

\begin{thebibliography}{10}
\providecommand{\url}[1]{#1}
\csname url@samestyle\endcsname
\providecommand{\newblock}{\relax}
\providecommand{\bibinfo}[2]{#2}
\providecommand{\BIBentrySTDinterwordspacing}{\spaceskip=0pt\relax}
\providecommand{\BIBentryALTinterwordstretchfactor}{4}
\providecommand{\BIBentryALTinterwordspacing}{\spaceskip=\fontdimen2\font plus
\BIBentryALTinterwordstretchfactor\fontdimen3\font minus \fontdimen4\font\relax}
\providecommand{\BIBforeignlanguage}[2]{{%
\expandafter\ifx\csname l@#1\endcsname\relax
\typeout{** WARNING: IEEEtran.bst: No hyphenation pattern has been}%
\typeout{** loaded for the language `#1'. Using the pattern for}%
\typeout{** the default language instead.}%
\else
\language=\csname l@#1\endcsname
\fi
#2}}
\providecommand{\BIBdecl}{\relax}
\BIBdecl
\renewcommand{\BIBentryALTinterwordstretchfactor}{4}

\bibitem{Jouaiti2019}
M.~Jouaiti and P.~Henaff, ``The sound of actuators: Disturbance in human - robot interactions?'' in \emph{Proc. ICDL-EpiRob '19}, 2019.

\bibitem{Allen2023}
A.~Allen, R.~Savery \emph{et~al.}, ``Consequential sounds and their effect on human robot interaction,'' in \emph{Sound and Robotics}, R.~Savery, Ed.\hskip 1em plus 0.5em minus 0.4em\relax CRC, 2023, ch.~6, pp. 101--127.

\bibitem{Langeveld2013}
L.~Langeveld, R.~van Egmond \emph{et~al.}, ``Product sound design: Intentional and consequential sounds,'' in \emph{Advances in Industrial Design Engineering}, D.~A. Coelho, Ed.\hskip 1em plus 0.5em minus 0.4em\relax InTech, 2013, ch.~3, pp. 47--73.

\bibitem{Wang2016}
L.~Wang and A.~Cavallaro, ``Ear in the sky: Ego-noise reduction for auditory micro aerial vehicles,'' in \emph{Proc. AVSS '16}.\hskip 1em plus 0.5em minus 0.4em\relax IEEE, 2016.

\bibitem{DePaivaVianna2015}
K.~M. D.~P. Vianna, M.~R.~A. Cardoso \emph{et~al.}, ``Noise pollution and annoyance: An urban soundscapes study,'' \emph{Noise and Health}, vol.~17, pp. 125--133, 2015.

\bibitem{Basner2014}
M.~Basner, W.~Babisch \emph{et~al.}, ``Auditory and non-auditory effects of noise on health,'' \emph{The Lancet}, vol. 383, pp. 1325--1332, 2014.

\bibitem{Jariwala2017}
H.~J. Jariwala, H.~S. Syed \emph{et~al.}, ``Noise pollution and human health: A review,'' in \emph{Proc. Noise and Air Pollution '17}, 2017.

\bibitem{Salvi2022}
R.~Salvi, G.~D. Chen \emph{et~al.}, ``Hyperacusis: Loudness intolerance, fear, annoyance and pain,'' \emph{Hearing Research}, vol. 426, 2022.

\bibitem{Henry2022}
J.~A. Henry, S.~M. Theodoroff \emph{et~al.}, ``Sound tolerance conditions (hyperacusis, misophonia, noise sensitivity, and phonophobia): Definitions and clinical management,'' \emph{American Journal of Audiology}, vol.~31, pp. 513--527, 9 2022.

\bibitem{He2023}
J.~L. He, Z.~J. Williams \emph{et~al.}, ``A working taxonomy for describing the sensory differences of autism,'' \emph{Molecular Autism}, vol.~14, 2023.

\bibitem{Yilmazyildiz2016}
S.~Yilmazyildiz, R.~Read \emph{et~al.}, ``Review of semantic-free utterances in social human-robot interaction,'' \emph{Int. J. Human-Computer Interaction}, vol.~32, pp. 63--85, 2016.

\bibitem{Zhang2023}
B.~J. Zhang and N.~T. Fitter, ``Nonverbal sound in human-robot interaction: A systematic review,'' \emph{THRI}, vol.~12, pp. 1--46, 2023.

\bibitem{Cha2018}
E.~Cha, N.~T. Fitter \emph{et~al.}, ``Effects of robot sound on auditory localization in human-robot collaboration,'' in \emph{Proc. HRI '18}, 2018.

\bibitem{Trovato2018}
G.~Trovato, R.~Paredes \emph{et~al.}, ``The sound or silence: Investigating the influence of robot noise on proxemics,'' in \emph{Proc. RO-MAN '18}, 2018.

\bibitem{Schute2007}
M.~Schütte, A.~Marks \emph{et~al.}, ``The development of the noise sensitivity questionnaire,'' \emph{Noise and Health}, vol.~9, pp. 15--24, 2007.

\bibitem{Moore2017}
D.~Moore, H.~Tennent \emph{et~al.}, ``Making noise intentional: A study of servo sound perception,'' in \emph{Proc. HRI '17}, 2017.

\bibitem{Wong2012}
P.~C. Wong, V.~Ciocca \emph{et~al.}, ``Effects of culture on musical pitch perception,'' \emph{PLoS ONE}, vol.~7, 2012.

\bibitem{Sneddon2003}
M.~Sneddon, K.~Pearsons \emph{et~al.}, ``Laboratory study of the noticeability and annoyance of low signal-to-noise ratio sounds,'' \emph{Noise Control Engineering Journal}, vol.~51, pp. 300--305, 2003.

\bibitem{Moore2018}
D.~Moore and W.~Ju, ``Sound as implicit influence on human-robot interactions,'' in \emph{Proc. HRI '18}, 2018.

\bibitem{Frederiksen2019}
M.~R. Frederiksen and K.~Stoey, ``Augmenting the audio-based expression modality of a non-affective robot,'' in \emph{Proc. ACII '19}, 2019.

\bibitem{Frederiksen2020}
M.~R. Frederiksen and K.~Stoy, ``Robots can defuse high-intensity conflict situations,'' in \emph{Proc. IROS '20}, 2020.

\bibitem{Song2017}
S.~Song and S.~Yamada, ``Expressing emotions through color, sound, and vibration with an appearance-constrained social robot,'' in \emph{Proc. HRI '17}, 2017.

\bibitem{Tennent2017}
H.~Tennent, D.~Moore \emph{et~al.}, ``Good vibrations: How consequential sounds affect perception of robotic arms,'' in \emph{Proc. RO-MAN '17}, 2017.

\bibitem{Zhang2021}
B.~J. Zhang, K.~Peterson \emph{et~al.}, ``Exploring consequential robot sound: Should we make robots quiet and kawaii-et?'' in \emph{Proc. IROS '21}, 2021.

\bibitem{Izui2020}
T.~Izui and G.~Venture, ``Correlation analysis for predictive models of robot user’s impression: A study on visual medium and mechanical noise,'' \emph{Int. Journal of Social Robotics}, vol.~12, pp. 425--439, 2020.

\bibitem{Frid2018}
E.~Frid, R.~Bresin \emph{et~al.}, ``Perception of mechanical sounds inherent to expressive gestures of a nao robot - implications for movement sonification of humanoids,'' in \emph{Proc. SMC '18: Sonic Crossings}, 2018.

\bibitem{Frid2022}
E.~Frid and R.~Bresin, ``Perceptual evaluation of blended sonification of mechanical robot sounds produced by emotionally expressive gestures: Augmenting consequential sounds to improve non-verbal robot communication,'' \emph{Int. Journal of Social Robotics}, 2022.

\bibitem{Tian2021}
L.~Tian, P.~Carreno-medrano \emph{et~al.}, ``Redesigning human-robot interaction in response to robot failures : a participatory design methodology,'' in \emph{Proc. CHI ’21}, 2021.

\bibitem{Wang2023}
Z.~Wang, Z.~Hu \emph{et~al.}, ``The effects of natural sounds and proxemic distances on the perception of a noisy domestic flying robot,'' \emph{THRI}, vol.~12, pp. 50:1--50:32, 2023.

\bibitem{Zhang2021b}
B.~J. Zhang, N.~Stargu \emph{et~al.}, ``Bringing wall-e out of the silver screen: Understanding how transformative robot sound affects human perception,'' in \emph{Proc. ICRA '21}, 2021.

\bibitem{Robinson2021}
F.~A. Robinson, M.~Velonaki \emph{et~al.}, ``Smooth operator: Tuning robot perception through artificial movement sound,'' in \emph{Proc. HRI '21}, 2021.

\bibitem{Bartneck2009}
C.~Bartneck, D.~Kulić \emph{et~al.}, ``Measurement instruments for the anthropomorphism, animacy, likeability, perceived intelligence, and perceived safety of robots,'' \emph{Int. Journal of Social Robotics}, vol.~1, pp. 71--81, 2009.

\bibitem{Schaefer2016}
K.~E. Schaefer, ``Measuring trust in human robot interactions: Development of the “trust perception scale-hri”,'' \emph{Robust Intelligence and Trust in Autonomous Systems}, pp. 191--218, 2016.

\bibitem{Fastl2007}
H.~Fastl and E.~Zwicker, ``Masking,'' in \emph{Psychoacoustics: Facts and Models}.\hskip 1em plus 0.5em minus 0.4em\relax Springer, 2007, ch.~4, pp. 61--110.

\bibitem{Hoffman2020}
G.~Hoffman and X.~Zhao, ``A primer for conducting experiments in human-robot interaction,'' \emph{ACM Transactions on Human-Robot Interaction}, vol.~10, 2020.

\bibitem{GPowerFaul2007}
F.~Faul, E.~Erdfelder \emph{et~al.}, ``G*power 3: A flexible statistical power analysis program for the social, behavioral, and biomedical sciences,'' \emph{Behavior Research Methods}, vol.~39, pp. 175--191, 2007.

\bibitem{Lieberman2022}
A.~Lieberman, J.~Schroeder \emph{et~al.}, ``A voice inside my head: The psychological and behavioral consequences of auditory technologies,'' \emph{Organizational Behavior and Human Decision Processes}, vol. 170, 2022.

\bibitem{Kallinen2007}
K.~Kallinen and N.~Ravaja, ``Comparing speakers versus headphones in listening to news from a computer - individual differences and psychophysiological responses,'' \emph{Computers in Human Behavior}, vol.~23, pp. 303--317, 2007.

\bibitem{Blumstein2016}
D.~T. Blumstein, ``Habituation and sensitization: new thoughts about old ideas,'' \emph{Animal Behaviour}, vol. 120, pp. 255--262, 2016.

\end{thebibliography}

\end{document}